\documentclass{article}

\usepackage[all]{nowidow}
\usepackage[utf8]{inputenc}
\usepackage[T1]{fontenc}
\usepackage{hyperref}
\usepackage{url}
\usepackage{times}

\usepackage{booktabs} 
\usepackage{graphicx} 
\usepackage{subcaption} 
\usepackage{amssymb}
\usepackage{amsmath}
\usepackage{bm}
\usepackage[autostyle=true]{csquotes} 
\usepackage{listings}
\usepackage{adjustbox}
\usepackage{multirow}
\usepackage{longtable}


\usepackage{tikz} 
\usetikzlibrary{positioning, chains, arrows.meta, shapes.geometric}

\usepackage{verbatim}

\usepackage{subcaption}
\usepackage{amsmath}
\usepackage{chngcntr}  

\usepackage[style=apa,
			citestyle=authoryear,
			sorting=nyt,
			sortcites=true,
			autopunct=true,
			babel=hyphen,
			hyperref=true,
			abbreviate=false,
			backref=false,
			dashed=false,
			maxcitenames=2, 
		    apamaxprtauth=999, 
			uniquename=false,
			uniquelist=false,
			backend=biber]{biblatex}

\title{A meta-analysis on the performance of machine-learning based language models for sentiment analysis}

\author{~
}

\author{
    Elena Rohde$^{1,\ast}$, Jonas Klingwort$^{2}$ and Christian Borgs$^{3}$\\
    \\
    \normalsize{$^{1}$ Institute of Sociology, Faculty of Social Sciences University of Duisburg-Essen,}\\ \normalsize{Lotharstr. 65, 47057 Duisburg, Germany} \\
    \normalsize{$^{2}$Department of Research \& Development, Statistics Netherlands (CBS), CBS-weg 11,}\\ \normalsize{PO Box 4481, 6401 CZ Heerlen, the Netherlands}\\
     \normalsize{$^{3}$IT.NRW -- Statistical Office of NRW, Mauerstr. 51, 40476  Düsseldorf, Germany} \\
    \normalsize{$^\ast$correspondence: elena.rohde@stud.uni-due.de}
}

\date{}

\begin{document} 

\maketitle 

\begin{abstract}
\noindent This paper presents a meta-analysis evaluating ML performance in sentiment analysis for Twitter data. The study aims to estimate the average performance, assess heterogeneity between and within studies, and analyze how study characteristics influence model performance. Using PRISMA guidelines, we searched academic databases and selected 195 trials from 20 studies with 12 study features. Overall accuracy, the most reported performance metric, was analyzed using double arcsine transformation and a three-level random effects model. The average overall accuracy of the AIC-optimized model was 0.80 $[0.76, 0.84]$. This paper provides two key insights: 1) Overall accuracy is widely used but often misleading due to its sensitivity to class imbalance and the number of sentiment classes, highlighting the need for normalization. 2) Standardized reporting of model performance, including reporting confusion matrices for independent test sets, is essential for reliable comparisons of ML classifiers across studies, which seems far from common practice.
\end{abstract}


\newpage
\section{Introduction}\label{sec: Introduction}


Social media is a valuable data source for social science research, particularly in analyzing public sentiment during events with considerable social impact \parencite{wang2021h}. However, the large volume of text data makes evaluation challenging. Sentiment analysis, using Natural Language Processing, extracts attitudes and emotions from text to classify content into categories like positive, negative, or neutral \parencite{govindarajan2022}. 


Sentiment analysis methods fall into lexicon-based and machine-learning approaches, with the latter preferred for social media due to higher accuracy \parencite{verma2022m, hartmann2019}. Machine learning strategies vary by algorithm and feature extraction, making overall performance evaluation challenging. This raises questions about algorithm effectiveness and the factors influencing variability. Identifying study characteristics and potential variability sources is crucial for setting realistic performance expectations \parencite{hartmann2023}.

This paper contributes to the literature by conducting a systematic literature review, followed by a meta-analysis and meta-regression, to explain the variation in the performance outcomes of machine learning algorithms in the context of social media data sentiment analysis. The results provide evidence of the factors contributing to the varying performance of different machine-learning algorithms in sentiment analysis.



\section{Research background}\label{sec:background}

Sentiment analysis is a method to analyze opinions, attitudes, and emotions towards entities expressed in text. The method was established in the early 1990s, extracting sentiment adjectives and analyzing viewpoints and metaphors \parencite{liu2015a}. In the early 2000s, sentiment analysis became more established and has been the subject of increasing research in recent years.



Given the growing interest in sentiment analysis, understanding performance factors is important. Key factors include training data size, balance, language, and text length. Larger datasets improve accuracy \parencite{hartmann2023, choi2017}, while unbalanced data yields higher accuracy than balanced datasets \parencite{hartmann2023}. English texts are classified more accurately than Dutch or French \parencite{boiy2009, hartmann2023}. Longer texts aid classification, especially with advanced models like BERT. Model and algorithm choices also impact performance \parencite{schuller2015, hartmann2023}.

Advanced techniques, including neural networks and ensemble methods like the extra tree classifier, tend to outperform traditional methods like support vector machines or decision trees in sentiment classification \parencite{rustam2021, hartmann2023, stine2019}. Furthermore, the achieved performance is also affected by the effectiveness of the feature extraction methods. 
Finally, fewer sentiment classes generally result in better model performance \parencite{bouazizi2019, hartmann2023}. To sum up, recent studies identify several key factors that affect classification quality, including the number of sentiment classes, text length and language, machine learning method, feature extraction technique, training dataset size, and sentiment class distribution within the training data. 


Few studies systematically analyze sentiment analysis performance factors. \textcite{hartmann2019} examines method performance across social media datasets, while the only meta-analysis by \textcite{hartmann2023} reports an average accuracy of 78\%, highlighting study-specific variability. Despite the growing number of publications, the lack of meta-analyses underscores the need for such studies to identify trends, generalize findings, and assess heterogeneity \parencite{boiy2009, hartmann2023}.

This paper aims to analyze how study designs affect the performance of machine learning algorithms in sentiment analysis and estimate the expected machine learning performance. In contrast to the existing meta-analysis from \textcite{hartmann2023}, the focus is on data from Twitter only. While \textcite{hartmann2023} compare lexicon approaches, traditional machine learning algorithms, and transfer learning models, this paper compares a more differentiated feature about machine learning algorithms. Furthermore, this paper uses additional features potentially affecting classification quality or potentially indicating the overall quality of the published paper, such as the labeling method, size of the majority class, or whether a confusion matrix was reported.

\section{Methods}\label{sec:methods}

A systematic literature review was conducted to identify relevant papers providing evidence for the research question (Section \ref{subsec:systematic}). The review is followed by a meta-analysis and meta-regression (Section \ref{subsec:meta1}).

\subsection{Systematic literature review and article selection}\label{subsec:systematic}


\subsubsection*{Inclusion criteria}

First, to adhere to our inclusion criteria, studies that perform sentiment analysis with machine learning techniques on Twitter data must report a quality metric: classification accuracy. Accuracy was chosen because it was reported in most of the identified studies. Second, the studies must have been peer-reviewed and published in journals, conference proceedings, or collective monographs. Third, the search was restricted to the last year, in which the social media platform Twitter had its original name (2022). As of July 2023, the platform's name changed to `X' \parencite{shah2023}. Although the change of platform policies might be an interesting feature, this decision was made to narrow down the scope of this meta-analysis. Fourth, only studies written in English are included. Fifth, by the definition of sentiment analysis in chapter \ref{sec:background}, only concepts representing an opinion, attitude, judgment, or emotion were considered sentiments. 

\subsubsection*{Identification}

\begin{figure}[t!]
    \centering    \includegraphics[width=0.6\textwidth]{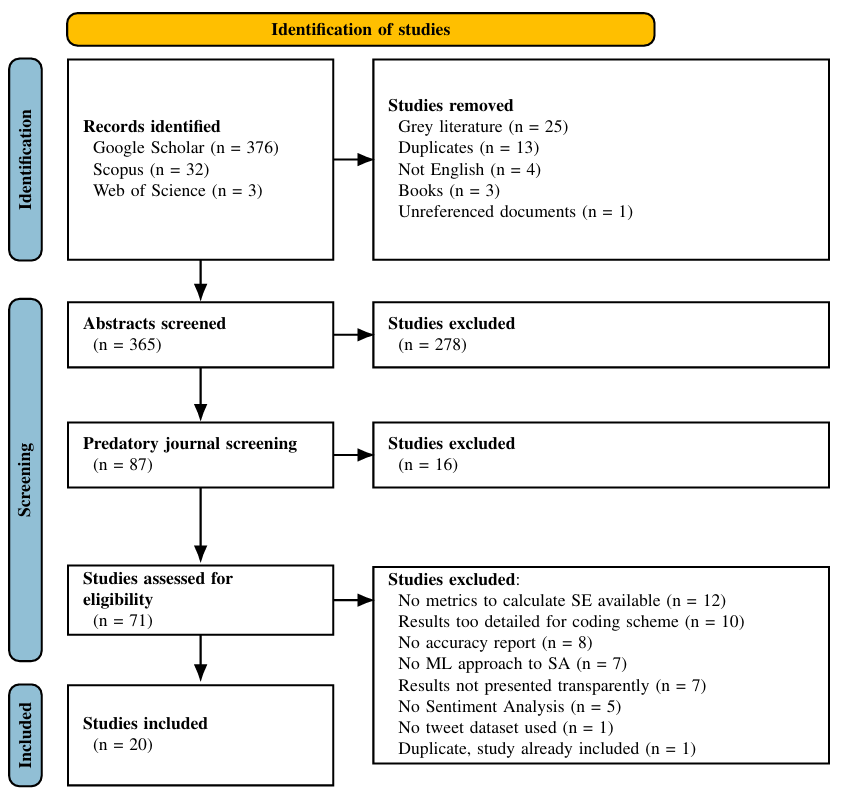}
\caption{Process of identifying records in a systematic literature search according to the PRISMA 2021 statement \parencite{page2021}.}
	\label{fig: PRISMA}
\end{figure}

Three large and relevant databases in the field of survey methodology and the social sciences \parencite{schnell2021} were considered: Scopus, Web of Science, and Google Scholar. The search term \enquote{\enquote{machine learning classifier} AND \enquote{sentiment analysis} AND \enquote{twitter}} was used. The process of identifying records according to the PRISMA guidelines is shown in Figure \ref{fig: PRISMA}. Finally, 411 records were identified. 
Gray literature, duplicates, non-english texts, books, and unreferenced documents were removed, leaving 365 records for abstract screening.

\subsubsection*{Screening: abstracts}

If the abstracts indicated a study using machine learning algorithms for sentiment analysis of Twitter data, they qualified for further review. For screening, the `metagear' R-library \parencite{lajeunesse2016} was used with R version 4.3.3 \parencite{rcoreteam2024}. General and specific keywords were chosen, and keywords that indicate inclusion and exclusion were selected. 
All authors conducted the screening independently to strengthen the quality and objectivity of this screening process. After screening, 278 studies were considered as not eligible.


\subsubsection*{Screening: predatory journals} 


The remaining 87 studies were checked for signs of predatory behavior. Beall's list and checklist \parencite{ross-white2019,beall2015} were used for this screening procedure. In total, 16 records were excluded, leaving 71 for a full-text eligibility assessment.

\subsubsection*{Screening: full texts}

The following study attributes should be included in the dataset \parencite{cuijpers2016}: study characteristics and features and information needed to calculate effect sizes. 51 records were excluded for the reasons shown in Figure \ref{fig: PRISMA}, leaving 20 records. These studies published the classification quality of multiple trials, resulting in $m=$ 195 reported overall accuracies distributed over the $h=20$ studies. Each study contributes a minimum of 2 and a maximum of 39 observations. 

\subsubsection*{Screening: study features}

\begin{figure}[h]
    \centering
    \includegraphics[width=.83\linewidth]{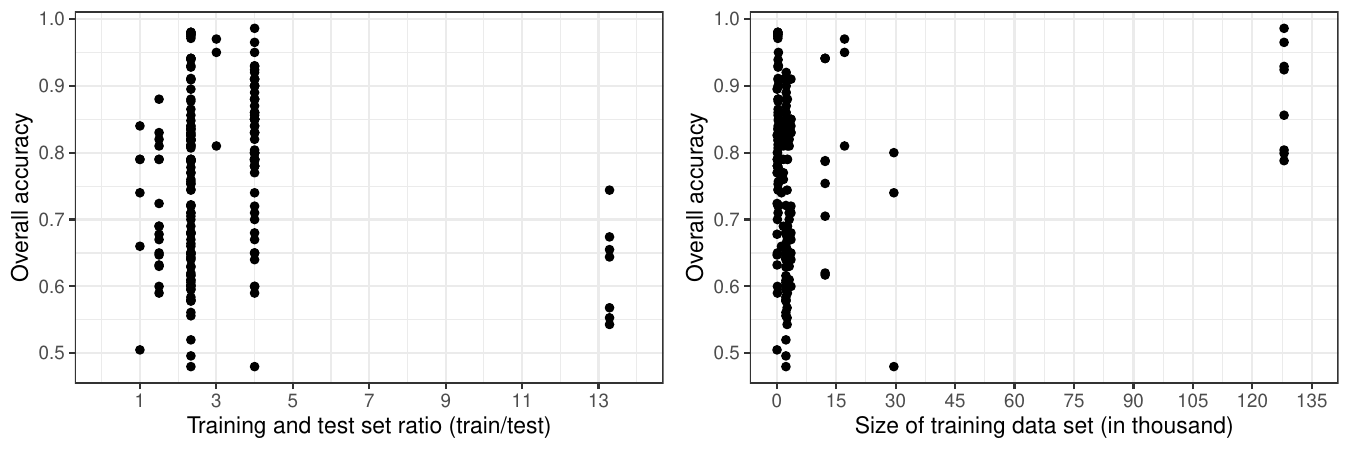}
    \caption{Distribution of overall accuracy by numeric feature.}    \label{fig:contplots}
\end{figure}

\begin{figure}[t]
    \centering
    \includegraphics[width=.83\linewidth]{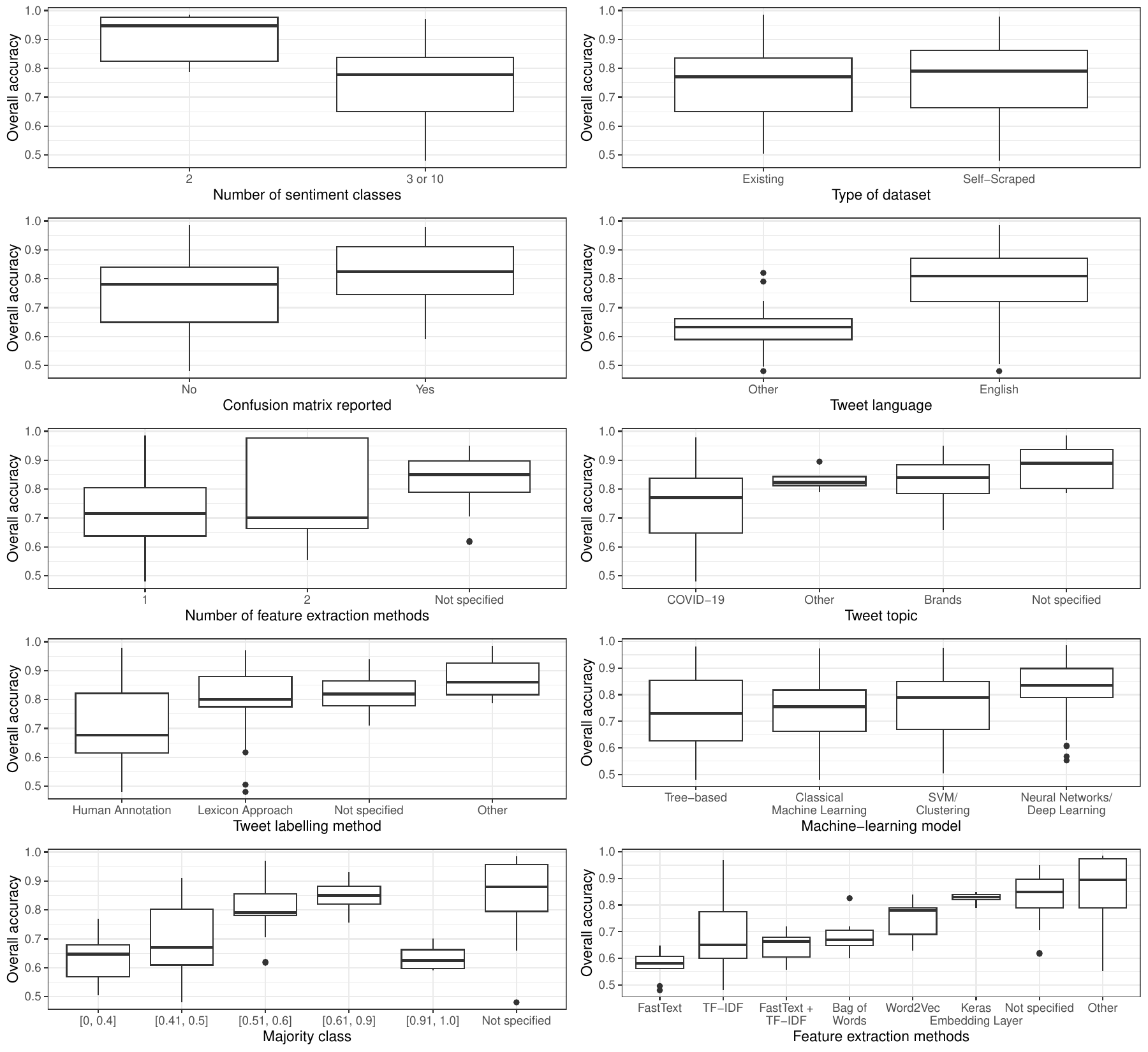}
    \caption{Distribution of overall accuracy by categorical feature.}    \label{fig:catplots}
\end{figure}

Twelve features were derived from the selected studies. Two could be used as numeric features (see Figure \ref{fig:contplots}). The feature `Size of the training data set (scaled)' was scaled by the factor $1e3$. Ten features were used as categorical (see Figure \ref{fig:catplots}). The feature `Number of sentiment classes' was categorized to avoid sparse categories. The `Number of feature extraction methods' and `Majority class' had to be categorized and could not be used as numeric features because some studies did not provide this information.

The feature `Training and test set ratio' is the size of training set divided by size of the test set. 
The category `Other' was introduced to group rare occasions to avoid sparse categories. For `Tweet language', the languages Nepali, Italian, and Tamil were grouped in `Other'. For `Tweet topic', topics on LGBTQ and railway infrastructure were grouped in `Other'. For the `Tweet labeling method,' a pre-labeled dataset or a combination of manual and machine-learning-based labeling was grouped in `Other.' For `Feature extraction methods', the method combining Bag of Words and TF-IDF, Count Vector, N-Grams, GloVe, and Bert Tokenizer was grouped in `Other'. If the information could not be retrieved, this was coded as `Not specified'. For the feature `Machine-learning model,' Logistic Regression, Naive Bayes, Multinomial Naive Bayes, and Genetic Algorithm were grouped as `Classical Machine-learning.' 

\newpage
\subsection{Statistical analysis}\label{subsec:meta1}

A meta-analysis of proportions is a statistical method used to synthesize and analyze results from multiple studies reporting proportions or rates, such as prevalence or incidence rates. Transferred to the current study, we aim to synthesize overall accuracies from various machine learning studies. 
For the statistical analyses, the R packages \texttt{metafor} \parencite{viechtbauer2010,viechtbauer2024} and \texttt{dmetar} \parencite{harrer2019} were used. The content and formulas in this section are based on the paper by \textcite{leach2025}.


\subsection*{Transforming proportions} 

This paper uses the Freeman-Tukey double arcsine transformation (FT) due to reported accuracies near 1.0, which cause variance instability under the logit transformation \parencite{freeman1950,warton2011,schwarzer2019,barendregt2013,wang2023a}. Other transformations, including arcsine square root, logit, and log, showed greater deviations than the double arcsine from normality (tested via Shapiro-Wilk). The double arcsine transformation was applied to the observed overall accuracy ($p_{ij}$) in trial $i$ of study $j$, resulting in the observed individual effect size, $\tilde{\theta}_{ij}$, and its sampling variance $v_{ij}$:

\begin{align}
    p_{ij} &= \frac{k_{ij}}{n_{ij}}, \\
    \tilde{\theta}_{ij} &= \text{FT}(p_{ij}) = \frac{\sin^{-1} \sqrt{\frac{k_{ij}}{n_{ij} + 1}} + \sin^{-1} \sqrt{\frac{k_{ij} + 1}{n_{ij} + 1}}}{2}, \\
    v_{ij} &= 1 / (4n_{ij} + 2).
\end{align}

\noindent with $k_{ij}$ being the number of correctly classified instances and $n_{ij}$ the number of instances in trial $i$ of study $j$. The transformed values are used to compute the summary effect size and standard error, with results back-transformed into raw proportions for easier reporting. For the back-transformation using the inverse population variance, we refer to \textcite{barendregt2013}.

\subsection*{Meta-analysis}

The independence among effect sizes is a key assumption in most meta-analytic models \parencite{cheung2014}. 
Hence, a multilevel random effect model was used to estimate the population effect size and heterogeneity between and within the studies, accounting for the dependency of individual effect sizes within each study $j$. A random-effects model has to be chosen when the studies are expected to have different underlying effects, often due to variability in study populations, interventions, or settings; it accounts for both within-study and between-study variation.

First, there is the individual level (1). Here, $\tilde{\theta}_{ij}$ is the observed individual effect size from trial $i$ in study $j$. This level is modeled as the (unobserved) true individual effect size $\theta_{ij}$ plus an error term $\epsilon_{ij}$. The error term is assumed to follow a normal distribution with a mean of 0 and a sampling variance of $v_{ij}$:

\begin{align}
    \tilde{\theta}_{ij} &= \theta_{ij} + \epsilon_{ij}, \nonumber \\
    \epsilon_{ij} &\sim N(0, v_{ij}).
\end{align}

\noindent Second, there is the study level (2). Here, the true individual effect size $\theta_{ij}$ for trial $i$ in study $j$ is expressed as the study effect size $\kappa_j$ from study $j$ and the within-study heterogeneity $\zeta_{ij}$. This heterogeneity is assumed to follow a normal distribution with a mean of 0 and a within-study variance of $\sigma^2_\zeta$:

\begin{align}
    \theta_{ij} &= \kappa_j + \zeta_{ij}, \nonumber \\
    \zeta_{ij} &\sim N(0, \sigma^2_\zeta).
\end{align}

\noindent Third, there is the population level (3). Here, the study effect size $\kappa_j$ in study $j$ is modeled as the population effect size, $\mu$, and between-study heterogeneity $\xi_j$. This heterogeneity is assumed to follow a normal distribution  with mean 0 and between-study variance $\sigma^2_\xi$:

\begin{align}
    \kappa_j &= \mu + \xi_j, \nonumber \\
    \xi_j &\sim N(0, \sigma^2_\xi).
\end{align}

\noindent The observed individual effect size is decomposed into the population effect size, the between-study heterogeneity, the within-study heterogeneity, and an error component: $\tilde{\theta}_{ij} = \mu + \xi_j + \zeta_{ij} + \epsilon_{ij}$. Its total variance is $\mathrm{Var}(\tilde{\theta}_{ij}) = \sigma^2_\xi + \sigma^2_\zeta + v_{ij}$.

The population effect size $\mu$ is a weighted sum of the observed individual effect sizes \parencite{konstantopoulos2011,cheung2014}. The weight considers the covariance between individual effect sizes within a study. The particular effect sizes in different studies are assumed to be independent, and individual effect sizes within a study share the same covariance. The restricted maximum likelihood (REML) is used to estimate the parameters.

The total variance is expressed as $\mathrm{Var}(\tilde{\theta}) = \sigma^2_\xi + \sigma^2_\zeta + \sigma^2_\epsilon$, where $\sigma^2_\epsilon$ represents a pooled estimate of the sampling variances $v_{ij}$ \parencite{higgins2002} to evaluate the relative contributions of the three variance components to the total variance.

\subsection*{Quantifying heterogeneity}

To assess potential heterogeneity, several metrics are used. First, $\sigma^2_\xi$ and $\sigma^2_\zeta$ quantify the variance of true effect sizes \parencite{harrer2022}, informing about variation within studies (level 2) and between studies (level 3) in a three-level meta-analysis. Second, Cochran's $Q$ \parencite{cochran1954} tests whether variation exceeds what is expected from sampling error. Third, the $I^2$ statistic \parencite{higgins2002} measures the proportion of variability not due to sampling error \parencite{harrer2022}. The metrics $Q$ and $I^2$ indicate how much of the observed variation exceeds the expected variation under homogeneity. 

\subsection*{Meta regression}\label{sec:metareg}

Meta-regression is a type of moderator analysis that attempts to explain the identified heterogeneity by examining the patterns of heterogeneity in the data \parencite{wang2023a,harrer2022}. Meta-regression can be described as the study-level equivalent of regression analysis in individual-level data, as study features are independent variables and the observed effect size is the dependent variable to be predicted \parencite{harrer2022}. The study features might explain the potentially identified heterogeneity between and within studies. For this purpose, the multilevel model is extended and becomes:

\begin{equation}
    \tilde{\theta}_{ij} = \mu + \bm{x}_j^\intercal \bm{\beta}_j + \xi_j + \zeta_{ij} + \epsilon_{ij},
\end{equation}

\noindent with $\bm{\beta}_j$ being a vector of $p$ regression coefficients of study study $j$ and $\bm{x}_j^\intercal$ denotes the transpose of a vector. The heterogeneity $\xi_j + \zeta_{ij}$ is now the variability across the true individual effect sizes $\theta_{ij}$ that is not explained by the features. Adding features to the model will reduce the variances at the population level ($\xi$) and study level ($\zeta$). The reduction is quantified by $R^2$:

\begin{align}
    R^2_\xi &= 1 - \frac{\sigma^2_\xi(X)}{\sigma^2_\xi(0)}, \\
    R^2_\zeta &= 1 - \frac{\sigma^2_\zeta(X)}{\sigma^2_\zeta(0)},
\end{align}

\noindent where $\sigma^2(X)$ is the variance according to the model with features and $\sigma^2(0)$ the variance according to the null model. Five models will be fitted; the first is the null model. Second, the full model uses all the collected features. Third, a model optimized towards the AIC; fourth, a model optimized towards the BIC; and fifth, a model optimized towards the RMSE will be shown. The null model serves as a baseline and will inform on the potential improvements of the other models. The complete model uses the twelve collected features. The AIC-optimized model balances fit and complexity. The BIC-optimized model has a stronger penalty for model complexity than AIC, reducing the risk of overfitting. The RMSE-optimized model does not penalize complexity, meaning a more complex model with lower RMSE might still be overfitting.

\section{Results}\label{sec:results}

In the following, the results of the meta-analysis and meta-regression are shown. This includes the review of the summary effect size, the quantification of heterogeneity, and the subsequent meta-regression.






\subsection{Three-level meta-analysis}

\begin{figure}[ht!]
	\centering
	\includegraphics[trim={2em 1em 2em 1em}, clip, width=\textwidth]{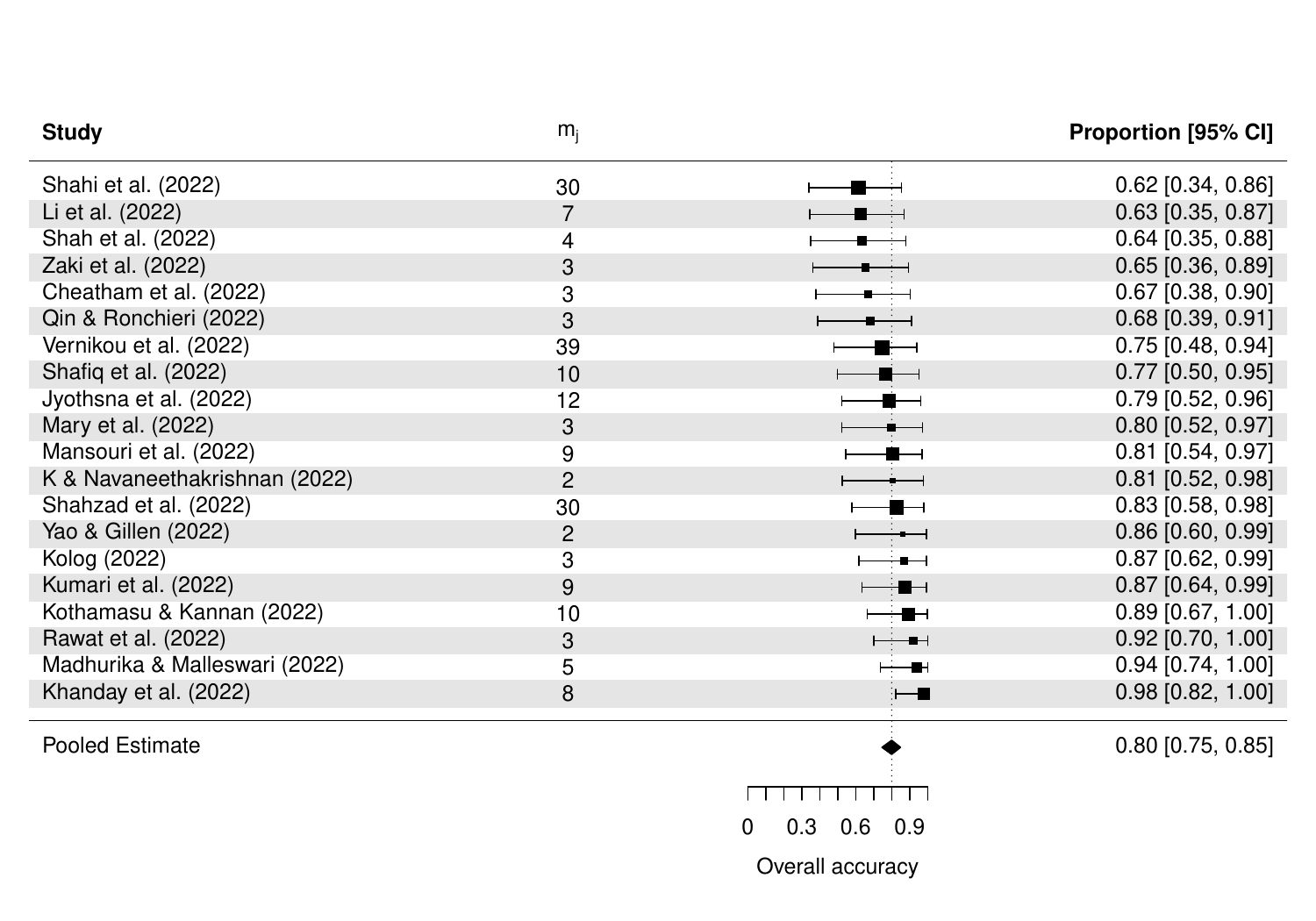}
    \caption{Forest plot of three-level meta-analysis.}
	\label{fig:forestplotagg}
\end{figure}

The forest plot in Figure \ref{fig:forestplotagg} shows the 20 individual studies, how many trials each study contributes, and their average summary effect size. The weighted average of the overall accuracy is estimated at 0.8 (95\% CI: 0.75;0.85). Cochran's Q-test yielded a significant result ($Q = 518042$, $p<0.001$), indicating the presence of heterogeneity beyond chance, i.e., sampling error. The variance of the true effect sizes indicates heterogeneity between and within studies (see Table \ref{tab:sumfitmodels}). The within-study variation attributable to the second level is higher, accounting for about 29\% of the data variation. Most of the variation in the data is due to between-study heterogeneity at level three, with about 71\%. All measures suggest that a large proportion of the heterogeneity in the data is not due to sampling error and that there is substantial heterogeneity between studies. These results justify the application of a meta-regression to attempt to explain the identified heterogeneity. 


\subsection{Three-level meta-regression}

Table \ref{tab:sumfitmodels} shows five fitted models. 

\begin{table}[htbp]
  \resizebox{\textwidth}{!}{%
    \centering
    \begin{tabular}{lrrrrrrrrrrr}
    \toprule
     Model &  $f$ & AIC & BIC & RMSE & $Q$ &  $\sigma^2_\xi$ ($h = 20$) & $\sigma^2_\zeta$ ($m = 195$) & $\hat{\mu}$ [$\hat{\mu}_l;\hat{\mu}_u$] & $R^2_\xi$ & $R^2_\zeta$\\
     \midrule
        AIC & 6 &  -358.92 & -332.99 &  0.12  & 178469 & 0.013 ($I^2_\xi 0.68$) & 0.006 ($I^2_\zeta = 0.32$) & 0.80 [0.76;0.84] & 0.34 & 0.23 \\
          BIC & 4 &  -354.39 &  -334.88 &  0.15 &  252323 & 0.020 ($I^2_\xi 0.76$) & 0.006 ($I^2_\zeta = 0.24$) & 0.80 [0.75;0.85] & - 0.01 & 0.23\\
     Null   & 1 & -322.00 & -312.20 & 0.16 & 518042 & 0.020 ($I^2_\xi 0.71$) & 0.008 ($I^2_\zeta = 0.29$) & 0.80 [0.75;0.85]\\
     Full   & 29 & -314.88 & -218.41 & 0.08 & 81363 & 0.014 ($I^2_\xi 0.70$) & 0.006 
 ($I^2_\zeta = 0.30$) & 0.80 [0.76;0.84] & 0.33 & 0.30\\
     RMSE & 29 & -314.88 &  -218.41 &  0.08  & 81363 & 0.014 ($I^2_\xi 0.70$) & 0.006 ($I^2_\zeta = 0.30$) & 0.80 [0.76;0.84] & 0.33 & 0.30\\
     \bottomrule
    \end{tabular}
    }
    \caption{Summary of fitted models. Rows are ordered by AIC.}
    \label{tab:sumfitmodels}
\end{table}

\noindent The number of selected features ($f$) varies from one in the null model to 29 in the full and RMSE model. Between-study variance  $\sigma^2_\xi$ is the highest in the null and BIC-optimized models, suggesting larger heterogeneity. Within-study variance $\sigma^2_\zeta$ remains relatively stable, with the AIC-optimized model having the largest proportion. The Full and RMSE models explain 33\% of between-study variance and 30\% of within-study variance. The AIC model performs slightly better for between-study variance and worse for within-study variance. The BIC model has a negative explained variance (only regarding the between-study variance), indicating poor performance in reducing heterogeneity. All models estimate $\hat{\mu} = 0.80$ with overlapping confidence intervals. The Full and RMSE models provide the best fit (lowest RMSE) and explain the most variance. The AIC optimized model performs explains slightly more between-study variance than the Full model. The BIC model performs poorly, with higher heterogeneity and negative explained variance. Note that the p-values for all $Q$ statistics indicated the presence of remaining heterogeneity not caused by sampling error.

The Tables \ref{table:fullmodel}--\ref{table:bicmodel} show the regression output from the full, AIC, and BIC models. Note that the RMSE model equals the full model; therefore, it is not shown. All tables show values based on the transformed proportions. In Table \ref{table:fullmodel}, significant predictors are found for the `Machine-learning model' (SVM/Clustering, Neural Networks/Deep Learning) and `Feature extraction method' (Keras Embedding Layer, Other methods). Both show positive effects. In this model, redundant predictors were dropped from the model. That means that some predictors are collinear or perfectly correlated. That was the case for the feature `Machine-learning model' and the `Not specified' category. Table \ref{table:aicmodel} shows the AIC-optimized model. The features `Machine-learning method', `Tweet language', and `Number of sentiment classes' were selected for this model. All predictors show significant effects except for Tree-based models. Table \ref{table:bicmodel} shows the BIC-optimized model. For this model, only the feature `Machine-learning method' was selected. All predictors show significant effects except for Tree-based models.

\begin{table}[t]
\resizebox{\textwidth}{!}{%
\begin{tabular}{lcccrr}
\toprule
& $\beta$ & SE & $p$ & 95\% CI \\
\midrule
Intercept &                                             1.0698 & 0.4962 & 0.2765 & [-5.2346; 7.3743] \\  
Train/test ratio &                   0.0107 & 0.0111 & 0.3345 & [-0.0111; 0.0326] \\     
Size of training data (/1000) &                             -0.0003 & 0.0070 & 0.9714 & [-0.0140; 0.0135] \\      
\textbf{Number of sentiment classes} (Ref: 2 classes)  &&&& \\
3 or 10 classes &                      -0.2635 & 0.2752 & 0.5138 & [-3.7607; 3.2336]  \\    
\textbf{Machine-learning model} (Ref: Classical machine learning)  &&&& \\
SVM/Clustering &                     0.0495 & 0.0181 & 0.0070 & [ 0.0137; 0.0853] \\  
Tree-based &                            0.0307 & 0.0164 & 0.0629 & [-0.0017; 0.0630] \\
Neural Networks/Deep Learning &       0.1063 & 0.0234 & <.0001 & [ 0.0600; 0.1525] \\
\textbf{Number of feature extraction methods} (Ref: 1 method)  &&&& \\
2 methods &                         -0.0589 & 0.0937 & 0.5300 & [-0.2439; 0.1260] \\     
Not specified &             0.0142 & 0.0630 & 0.8224 & [-0.1102; 0.1385] \\
\textbf{Feature extraction method} (Ref: TF-IDF) &&&& \\
FastText &                -0.0525 & 0.0321 & 0.1030 & [-0.1158; 0.0107] \\
FastText + TF-IDF &       0.0850 & 0.0989 & 0.3910 & [-0.1102; 0.2802] \\     
Bag of Words &           -0.0060 & 0.0327 & 0.8551 & [-0.0705; 0.0585] \\     
Word2Vec &                -0.0543 & 0.0375 & 0.1495 & [-0.1284; 0.0198] \\     
Keras Embedding Layer&    0.0977 & 0.0352 & 0.0061 & [ 0.0283; 0.1672] \\ 
Other&                     0.0671 & 0.0309 & 0.0311 & [ 0.0062; 0.1281] \\
\textbf{Tweet language} (Ref: English)  &&&& \\
Other &                              -0.0998 & 0.1069 & 0.5218 & [-1.4577; 1.2581]      \\
\textbf{Tweet labeling method} (Ref: Human annotation)  &&&& \\
Lexicon Approach &                  0.1448 & 0.1559 & 0.5233 & [-1.8356; 2.1252] \\     
Other &                             0.2415 & 0.2269 & 0.4801 & [-2.6417; 3.1248] \\    
Not specified &                    -0.1048 & 0.1707 & 0.5401 & [-0.4418; 0.2322] \\ 
\textbf{Majority class} (Ref: \text{[0; 0.4]})  &&&& \\
\text{[0.41, 0.5]}  &                       0.1766 & 0.1436 & 0.4346 & [-1.6479; 2.0011] \\     
\text{[0.51, 0.6]}  &                       0.1973 & 0.1331 & 0.1400 & [-0.0654; 0.4601] \\   
\text{[0.61, 0.9]}  &                       0.1574 & 0.1224 & 0.2000 & [-0.0841; 0.3990] \\     
\text{[0.91, 1.0]}  &                      -0.1557 & 0.1734 & 0.5342 & [-2.3589; 2.0475] \\     
Not specified &                      0.0916 & 0.2009 & 0.7277 & [-2.4608; 2.6440] \\  
\textbf{Tweet topic} (Ref: Brands) &&&& \\
COVID-19 &                            -0.0910 & 0.2083 & 0.7378 & [-2.7380; 2.5560]    \\  
Not specified &                       -0.3214 & 1.2687 & 0.8420 & [-16.4412; 15.7984]      \\
Other &                               -0.2186 & 0.4499 & 0.7121 & [-5.9349; 5.4977]      \\
\textbf{Type of dataset} (Ref: Existing)  &&&& \\
Self-Scraped         &               0.0428 & 0.1005 & 0.6704 & [-0.1555; 0.2412]  \\ 
\textbf{Confusion matrix reported} (Ref: No) &&&& \\
Yes                &                   0.2471 & 0.1672 & 0.3788 & [-1.8779; 2.3720] \\ 
\bottomrule
\end{tabular}
}
\caption{Meta-regression results of full model.}
\label{table:fullmodel} 
\end{table}

Overall, the machine-learning method is the most consistent predictor in all three models. Neural Networks/Deep Learning consistently show the strongest positive effect across all models. SVM/Clustering also indicates a positive impact, though weaker than Neural Networks. Tree-based models have a small positive effect, which is not statistically significant. Tweet language matters in the AIC model but not in the full model. The number of sentiment classes negatively affects the AIC model but is insignificant in the full model. Feature extraction methods show some impact in the full model but are removed in the AIC and BIC models. Many predictors (train/test ratio, dataset type, confusion matrix reporting, tweet labeling method, and majority class) are insignificant in any model, suggesting they do not strongly influence the outcome.  

\begin{table}[h]
\resizebox{\textwidth}{!}{%
\begin{tabular}{lcccr}
\toprule
& $\beta$ & SE & $p$ & 95\% CI \\
\midrule
Intercept &  1.1994 & 0.0580 &  <.0001  & [1.0750; 1.3239]  \\
\textbf{Machine learning method} (Ref: Classical machine learning) &&&& \\
SVM/Clustering & 0.0521 & 0.0185 &  0.0054 &  [0.0156; 0.0886]\\
Tree-based & 0.0268 & 0.0170 &  0.1157 & [-0.0067; 0.0604] \\      
Neural Networks/Deep Learning  &  0.1344 & 0.0195 &  <.0001  & [0.0958; 0.1729] \\
\textbf{Tweet language} (Ref: English)  &&&& \\
Other & -0.1825 & 0.0765 &  0.0317 & [-0.3465; -0.0185] \\
\textbf{Number of sentiment classes} (Ref: 2 classes)  &&&& \\
3 or 10 classes & -0.1559 & 0.0635 &  0.0277 & [-0.2921; -0.0197]\\
\bottomrule
\end{tabular}
}
\caption{Meta-regression results of AIC-optimized model.}
\label{table:aicmodel} 
\end{table}

\begin{table}[h]
\resizebox{\textwidth}{!}{%
\begin{tabular}{lcccr}
\toprule
& $\beta$ & SE & $p$ & 95\% CI \\
\midrule
Intercept & 1.0540 & 0.0346 & <.0001  & [0.9807; 1.1273]  \\
\textbf{Machine Learning Method}  (Ref: Classical machine learning)   &&&& \\
SVM/Clustering & 0.0514  & 0.0185 &   0.0061  & [0.0149;  0.0880] \\
Tree-based    &  0.0263  &0.0170  & 0.1237 & [-0.0073;  0.0599]      \\
Neural Networks/Deep Learning   & 0.1349  &0.0196 &  <.0001&   [0.0963;  0.1735] \\
\bottomrule
\end{tabular}
}
\caption{Meta-regression results of BIC-optimized model.}
\label{table:bicmodel} 
\end{table}

\section{Discussion}\label{sec: Discussion}

This meta-analysis evaluated the performance of language models for sentiment analysis using Twitter data. The study estimated the average performance, evaluated the level of heterogeneity between and within studies, and assessed which study features explain model performance. The overall population accuracy was estimated at $0.80~[0.75, 0.85]$ without features -- and at $0.80~[0.76, 0.84]$ when optimized for AIC and considering the machine-learning method used, the Tweet language, and the number of sentiment classes as features. The results demonstrate considerable heterogeneity between studies that the features in the AIC-optimized model could partially explain. Only the machine-learning method used remained a significant feature when optimizing for BIC. This result aligns with recent research by \textcite{hartmann2023}, who reported an average accuracy of about 78\% across all machine learning methods based on 217 publications containing 1,169 trials.

The key takeaway of this study is that deep learning consistently improves performance. SVM/Clustering performs better than classical machine learning but is less performant than deep learning. Tree-based models might be beneficial, but the evidence is weak. Tweet language (non-English) may reduce performance, but the effect is inconsistent. Using 3 or 10 sentiment classes instead of 2 negatively impacts performance. Feature extraction methods may influence performance, but this effect is inconsistent across models.

\subsection*{Limitations and future research}

First, we would like to discuss the more general issue of reporting (only) overall accuracy in sentiment analysis. This metric was chosen for this study as it was the most commonly reported. While common and intuitive, this metric has limitations, particularly with class imbalance and varying sentiment class counts. A model can achieve high accuracy by predicting the majority class, even if it misclassifies minority classes. Binary classification typically results in higher accuracy than three-class models due to increased difficulty in distinguishing sentiments. Key performance metrics like precision, recall, and F1-score, which are crucial for handling imbalanced data are rarely reported. We encourage reporting these metrics per class, along with a confusion matrix, as 77\% of the included trials in this study did not. 

Second, for some features, the categories `Not specified' and `Other' scored high in overall accuracy. It would be interesting to have more information, especially for the `Not specified' option. This finding suggests there is room for improvement and is a sign of poor reporting habits.

Third, no large language models were included because they started to gain public attention only after the reference period of this study. Future studies should study whether LLMs outperform the previously existing methods used in this study. 

Fourth, the systematic review considered English records published in 2022, which may not provide a complete picture of the available literature. The screening process involved subjective decisions, which, although documented and discussed with independent reviewers, may have influenced the choice of included records. 

Fifth, we want to reflect on the tweet labeling method. Annotating the raw data using methods other than manual human annotation may be problematic. The regression results show increased accuracy scores when lexicon approaches are used instead of human annotation. Given that sentiment lexicons are known to have lower accuracy than human annotation, this result indicates possibly a falsely inflated classification quality and should, therefore, be interpreted cautiously and considered in future research.

Future meta-analyses should build upon the current study and continue to explain and further understand the variation in the performance of machine learning algorithms in sentiment analysis. Therefore, it would be essential to include additional years of literature to gather more data points and add informational value in explaining the variation in classification performance. 

\newpage
\section{Conclusion}\label{sec:conclusion}

Monitoring and implementing sentiment analyses are considered practical and important for understanding public opinion. For this purpose, machine learning is a widely adopted and commonly used tool. While the overall accuracy of machine learning models applied to sentiment analysis is often considered good, it remains a popular yet inadequate quality metric for comparing studies. Overall accuracy is highly sensitive to class imbalance, and different studies vary in the degree of class imbalance. Relying solely on overall accuracy as a quality metric can create an overly optimistic impression of model performance. Therefore, reporting additional quality metrics, the confusion matrix, and normalizing them to ensure a more reliable and meaningful evaluation of sentiment analysis models is advisable.

\section*{Acknowledgments}

The views expressed in this report are those of the authors and do not necessarily correspond to the policies of Statistics Netherlands.\\

\noindent The data and code used in this study are available upon request. 

%
%
%




\section*{Reference list of included studies}

\begin{refsection}

\noindent The reference list below contains the studies included in the meta-analysis and meta regression.

\nocite{cheatham2022c, jalani2022c, k2022, khanday2022b, kolog2022c, kothamasu2022b, kumari2022l, li2022c, madhurika2022b, mansouri2022b, mary2022b, qin2022c, r2022c, rawat2022b, shah2022j, shahi2022c, shahzad2022c, vernikou2022b, yao2023, zaki2022}

\printbibliography[heading=subbibliography]

\end{refsection}

\newpage

\printbibliography

\newpage

\end{document}